\documentclass{article} 
\usepackage{iclr2027_conference,times}

\usepackage{amsmath,amsfonts,bm}

\def\eqref#1{equation~\ref{#1}}

\def\1{\bm{1}}

\DeclareMathAlphabet{\mathsfit}{\encodingdefault}{\sfdefault}{m}{sl}
\SetMathAlphabet{\mathsfit}{bold}{\encodingdefault}{\sfdefault}{bx}{n}

\usepackage{xcolor}
\usepackage[
colorlinks=true,
citecolor=blue!60!black,
linkcolor=black,
urlcolor=blue!60!black
]{hyperref}
\usepackage{url}
\usepackage{booktabs}
\usepackage{multirow}
\usepackage{makecell}
\usepackage{arydshln}
\usepackage{colortbl}
\usepackage{subcaption}
\usepackage{graphicx}
\usepackage{wrapfig}
\usepackage{booktabs}
\graphicspath{{./figure/}}

\definecolor{bestcolor}{RGB}{247,229,233}
\definecolor{secondcolor}{RGB}{201,223,243}

\newlength{\scorewidth}
\title{UltraMatch: Transport Path Routing for Ultra-Fast and Memory-Efficient Image Matching}

\author{%
Jiajun Le$^{1}$ \quad
Yifan Lu$^{1}$ \quad
Zizhuo Li$^{1}$ \quad
Lei Cao$^{1,2}$ \quad
Junjun Jiang$^{3}$ \quad
Jiayi Ma$^{1,4}$\thanks{Corresponding author.} \\[3pt]
$^{1}$Electronic Information School, Wuhan University, China \\
$^{2}$Xiaomi Corporation, China \\
$^{3}$School of Computer Science and Technology, Harbin Institute of Technology, China \\
$^{4}$School of Robotics, Wuhan University, China \\[3pt]
\small\texttt{\{jiajunle01,jyma2010\}@gmail.com, jiangjunjun@hit.edu.cn} \\
\small\texttt{\{lyf048,zizhuo\_li,whu.caolei\}@whu.edu.cn}
}

\iclrfinalcopy 
\begin{document}

\maketitle
\fancyhead{}
\lhead{Under review as a conference paper at ICLR 2027}
\renewcommand{\headrulewidth}{0.4pt}

\begin{abstract}
Despite recent advances in accuracy and efficiency, coarse matching remains an indispensable yet costly stage in existing semi-dense matchers due to dense token-level matching.
We present UltraMatch, an ultra-efficient and scalable semi-dense matching framework that bypasses the quadratic computation and memory cost of dense token-level matching by routing only a small fraction of candidate matching paths. At its core, a lightweight Transport Path Router operates on coarse block representations to rank candidate target blocks for each source block and retain only a small set, restricting subsequent token-level matching to the selected paths and avoiding the construction of the full token-to-token matching matrix. We further design a sparse global Dual-Softmax that performs matching only over the routed block candidates while retaining global competition across the sparse matching space. Beyond matching acceleration, UltraMatch employs deployment-oriented structural reparameterization for feature extraction and a tiny fine matching head with shared parameters, further reducing inference cost and memory consumption. UltraMatch achieves competitive accuracy among semi-dense matchers, while running 1.67$\times$ faster than SuperPoint+LightGlue with only 0.44 GiB peak inference memory.
Its scalability enables inference at up to 6K resolution on a single RTX 3090, whereas existing semi-dense matchers run out of memory before reaching 2K. Our routing strategy is also transferable, delivering about 2$\times$ end-to-end speedup in EDM and ELoFTR without accuracy loss. The project repository is available at
\url{https://github.com/JiajunLe/UltraMatch}.
\end{abstract}

\section{Introduction}
Image feature matching is a fundamental problem in computer vision and plays a central role in structure from motion (SfM)~\citep{schonberger2016sfm,he2024detectorfreesfm}, simultaneous localization and mapping (SLAM)~\citep{murartal2015orbslam,campos2021orbslam3}, and visual localization~\citep{sarlin2021pixloc}. Traditional pipelines typically detect sparse keypoints, describe them with local descriptors~\citep{lowe2004sift}, and establish correspondences according to descriptor similarity. With the development of deep learning, learned approaches have emerged for feature detection, description \citep{yi2016lift,potje2024xfeat}, and matching ~\citep{sarlin2020superglue,shi2022clustergnn}, substantially improving the robustness of local match estimation. More recently, detector-free methods~\citep{rocco2020sparsenchnet,sun2021loftr,huang2023adamatcher} directly match dense feature grids and refine selected grid correspondences to precise image coordinates, producing semi-dense matches with broader coverage and achieving strong performance across a wide range of geometric tasks.

\begin{figure}[t]
	\begin{subfigure}[b]{0.50\textwidth}
		\raggedright
		\includegraphics[width=\linewidth]{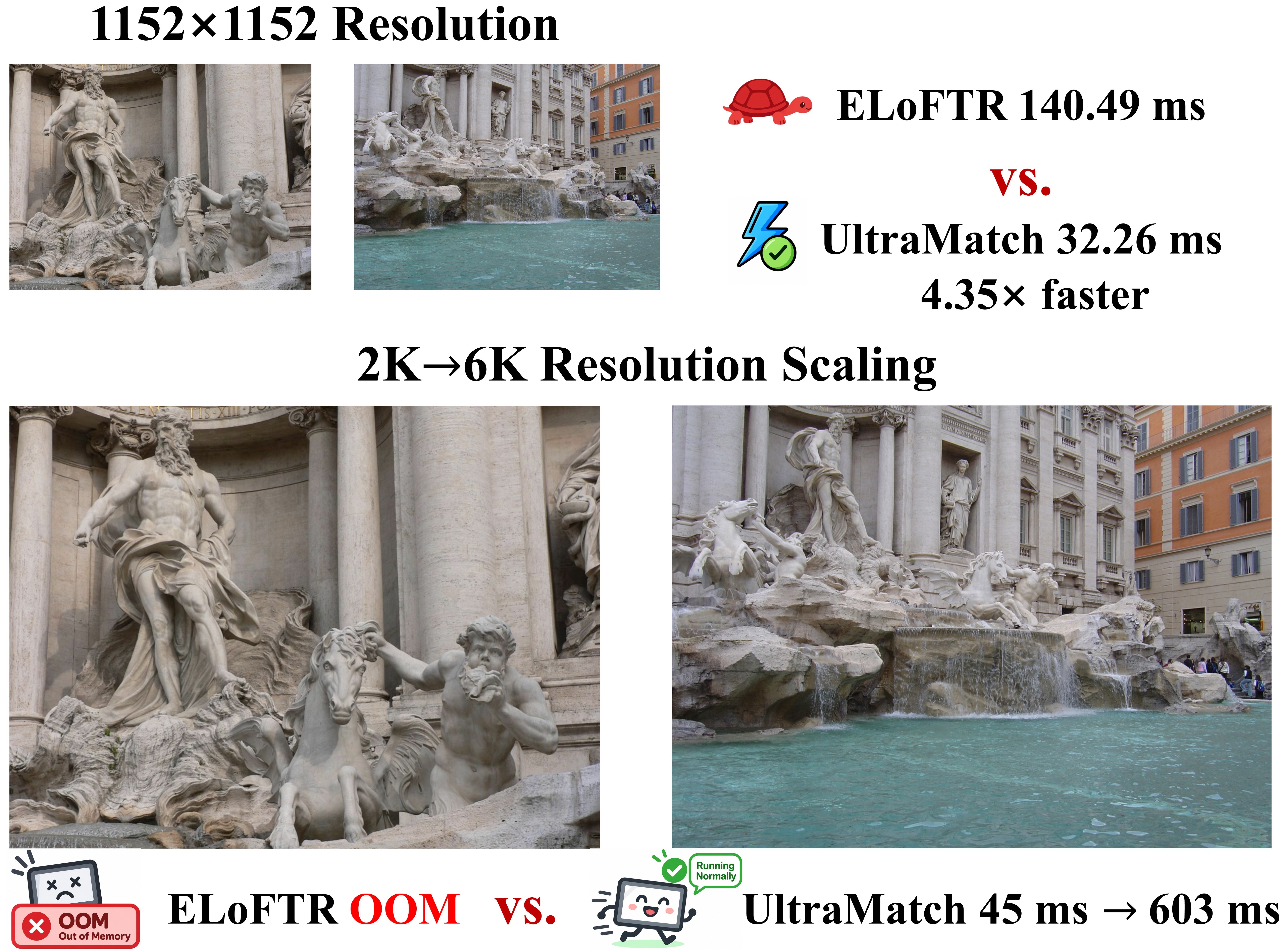}
		\caption{Resolution scalability.}
		\label{fig:intro_scalability}
	\end{subfigure}
	\hfill
	\begin{subfigure}[b]{0.45\textwidth}
		\raggedleft
		\includegraphics[width=\linewidth]{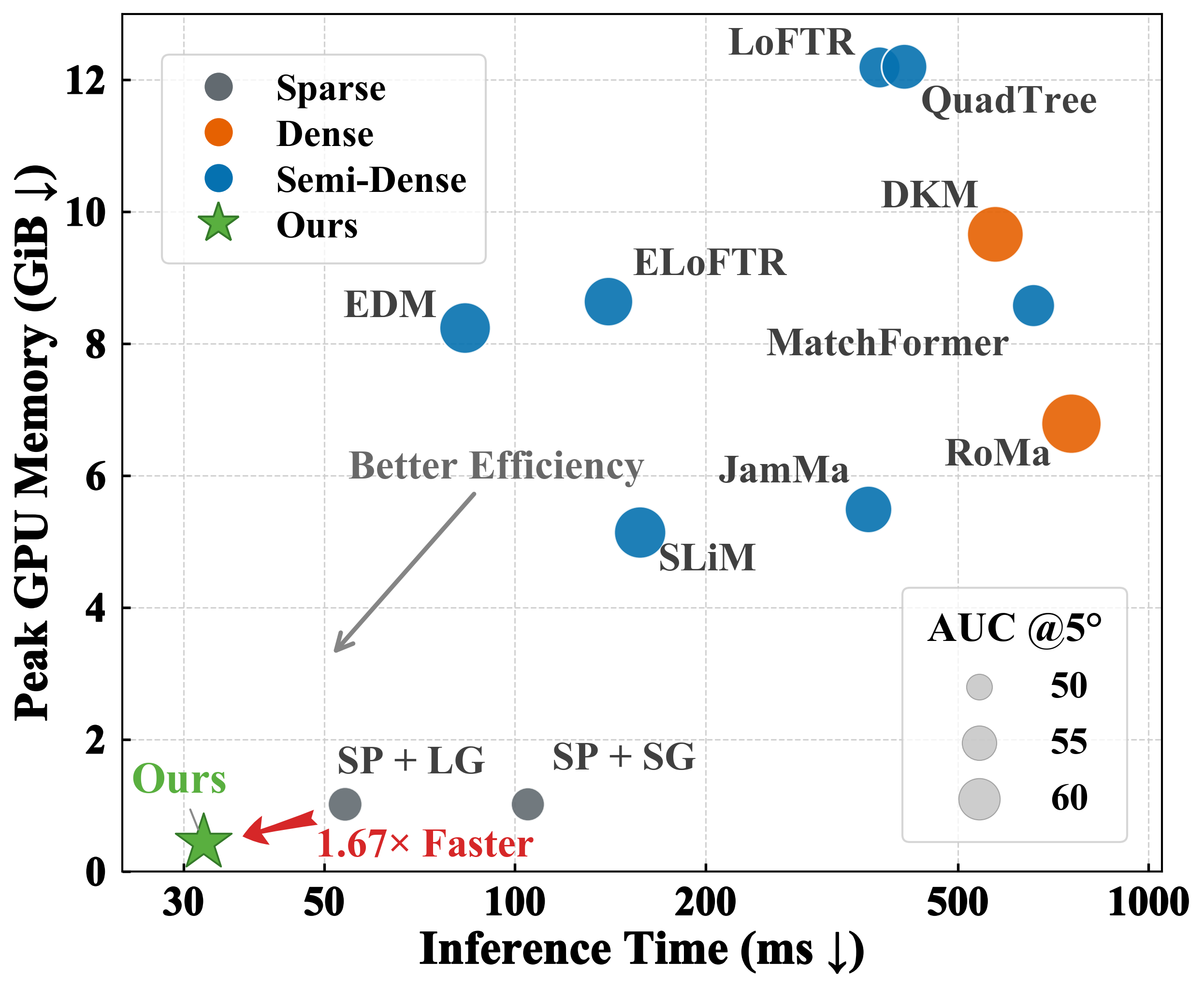}
		\caption{Efficiency trade-off.}
		\label{fig:efficiency_tradeoff}
	\end{subfigure}
	\vspace{-6pt}
	\caption{
		\textbf{Scalability and efficiency of UltraMatch.}
		(a) Runtime comparison across increasing input resolutions.
		(b) Efficiency trade-off in latency and memory consumption on MegaDepth-1500,
		where bubble size indicates AUC@5$^\circ$.
		All measurements are conducted on a single RTX 3090 GPU.
	}
	\label{intro_efficiency}
	\vspace{-13pt}
\end{figure}

Despite the strong progress in matching accuracy, current semi-dense matchers still incur substantial computation and memory overhead. Existing methods~\citep{chen2024ecomatcher,li2025edm} have improved efficiency across different stages of the pipeline, including feature extraction, feature interaction, coarse matching, and fine refinement. However, the coarse matching stage still commonly evaluates dense pairwise similarities between source and target tokens, resulting in quadratic computation and memory complexity with respect to the number of tokens. As the input resolution increases, the rapidly growing matching matrix leads to substantial latency and memory consumption, limiting real-time deployment and scalability. In practice, even recent efficient semi-dense matchers quickly approach the memory capacity of a modern GPU as the input resolution increases.

Although dense token matching compares each source token against the entire set of target tokens, each source token can eventually have at most one valid match. Constructing the complete pairwise matrix therefore spends substantial computation and memory on a large number of unnecessary candidates. Inspired by the multiscale optimal transport principle~\citep{schmitzer2016sparse} of progressively restricting candidate transport paths, dense matching can be represented as a complete bipartite graph between source and target tokens, which can then be reduced to a small subset of candidate edges before token level matching. In this way, expensive matching is performed only within a substantially smaller search space.

Building on this analysis, we propose UltraMatch, an ultra-efficient and scalable semi-dense matching framework that reduces dense matching computation through Transport Path Routing. The router operates on block-level representations that contain rich correspondence cues, allowing it to identify promising target blocks for each source block. Only the selected block candidates are then expanded to the token level, substantially reducing the number of pairwise similarities that need to be evaluated. Since block routing may miss valid matches near block boundaries, we extend each routed block by a one token margin before matching, introducing only limited additional computation. We further develop a sparse global Dual-Softmax that preserves competition among all routed candidates instead of performing matching independently within individual blocks. This allows UltraMatch to retain the matching performance of dense Dual-Softmax while operating on a much smaller candidate space.
To further improve efficiency, we also optimize feature extraction and fine refinement. For feature extraction, we employ structural reparameterization to increase representation flexibility during training while retaining a compact single branch structure for inference. For refinement, we design a lightweight fine matching head with shared query and reference encoders, avoiding duplicated branch-specific parameters while keeping subpixel refinement efficient.

Empirically, UltraMatch maintains competitive geometric accuracy while substantially improving efficiency and scalability, as shown in Fig.~\ref{intro_efficiency}. On MegaDepth~\citep{li2018megadepth}, it runs $1.67\times$ faster than the sparse SuperPoint+LightGlue~\citep{detone2018superpoint,lindenberger2023lightglue} pipeline and $4.35\times$ faster than ELoFTR~\citep{wang2024efficientloftr}, bringing semi-dense matching into the real-time regime. It scales to 6K on a single RTX 3090, while existing semi-dense matchers run out of memory below 2K. Moreover, applying Transport Path Routing to JamMa~\citep{lu2025jamma}, ELoFTR and EDM~\citep{li2025edm} reduces their end-to-end latency by approximately 50$\%$ without accuracy degradation. Our contributions are summarized as follows:
\begin{itemize}
	
	\item We propose UltraMatch, an ultra-efficient semi-dense matching framework centered on Transport Path Routing. The proposed router selects a small set of candidate block routes before token-level matching, while sparse global Dual-Softmax performs matching only over the routed candidates with global competition preserved, substantially reducing dense pairwise matching computation.
	
	\item We adopt structural reparameterization for efficient feature extraction and design a compact fine matching head with shared parameters for subpixel refinement, further improving the efficiency of the overall matching pipeline.

	\item Extensive experiments demonstrate competitive geometric accuracy, ultra-low latency, and strong memory scalability. Transport Path Routing also transfers effectively across semi-dense matchers, providing substantial acceleration with minimal accuracy loss.
	
\end{itemize}

\section{Related Work}
\subsection{Deep Local Feature Matching}
Deep local feature matching has evolved from sparse keypoint-based matching to dense and semi-dense correspondence estimation.
Sparse methods typically rely on keypoint detection, description~\citep{revaud2019r2d2,detone2018superpoint,tyszkiewicz2020disk,zhao2023aliked}, and matching~\citep{sarlin2020superglue,xue2023imp,jiang2024omniglue}. Among them, SuperPoint jointly learns detection and description, while SuperGlue employs self- and cross-attention to model correlations between sparse local features.
Dense methods~\citep{rocco2018ncnet,jiang2021cotr,ni2023pats,edstedt2023dkm,edstedt2024roma} instead estimate pixel-level correspondences, with DKM modeling them probabilistically and RoMa leveraging pretrained visual representations.
Semi-dense matching provides rich coverage without exhaustive pixel-level estimation~\citep{zhou2021patch2pix,chen2022aspanformer,sun2021loftr}. LoFTR establishes a Transformer-based coarse-to-fine paradigm, while MatchFormer~\citep{wang2022matchformer} interleaves feature extraction and feature interaction. 
More recently, HomoMatcher~\citep{wang2025homomatcher} incorporates homography estimation into fine-level refinement, and CoMatch~\citep{li2025comatch} introduces covisibility-aware interaction and bilateral subpixel refinement. Together, these methods have substantially advanced matching accuracy and robustness.

\subsection{Efficient Deep Feature Matching}
Despite substantial progress in matching accuracy, feature interaction and dense pairwise matching remain computational bottlenecks, especially at high resolutions.
For sparse matching, SGMNet~\citep{chen2021sgmnet} propagates information through reliable seed matches, while LightGlue~\citep{lindenberger2023lightglue} adaptively adjusts network depth and prunes unnecessary keypoints.
For dense matching, ArgMatch~\citep{deng2025argmatch} selectively allocates refinement to informative regions.
More efforts have recently focused on accelerating semi-dense matching. QuadTree~\citep{tang2022quadtree} organizes feature interaction hierarchically to reduce attention computation, while ELoFTR~\citep{wang2024efficientloftr} introduces aggregated attention and efficient correlation refinement to reduce the cost of the  pipeline. TopicFM+~\citep{giang2024topicfmplus} uses compact topic-based interaction, and ETO~\citep{ni2024eto} organizes multiple homography hypotheses to simplify correspondence estimation. More recently, JamMa~\citep{lu2025jamma} and SLiM~\citep{choo2026slim} adopt Mamba for lightweight feature interaction, while EDM~\citep{li2025edm} improves efficiency throughout the matching pipeline. Despite these advances, dense token-level matching still evaluates numerous pairwise candidates, causing substantial latency and memory overhead. UltraMatch addresses this remaining bottleneck by routing only a small fraction of candidate matching paths, reducing both matching computation and memory consumption, especially at high resolutions.

\section{Method}
\begin{figure}[t]
	\centering
	\includegraphics[width=\linewidth]{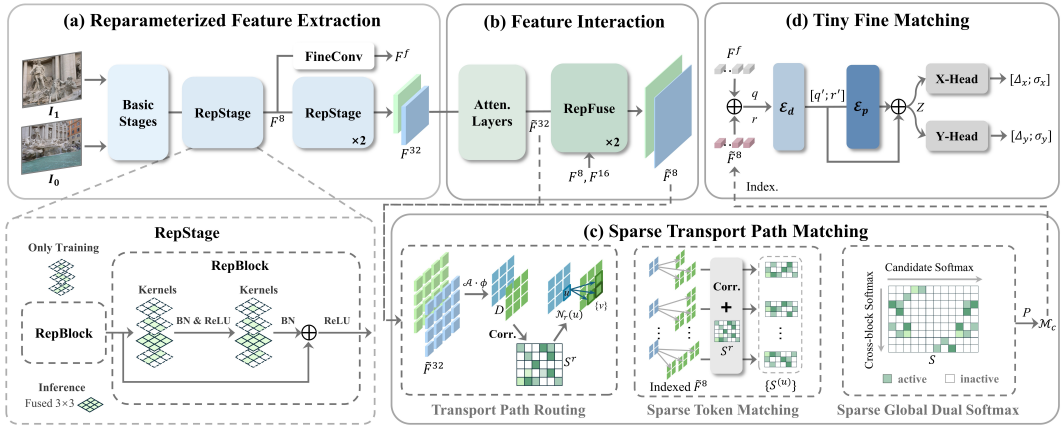}
	\caption{
		\textbf{Overview of UltraMatch.}
		(a) A lightweight backbone extracts multi-scale features, with reparameterizable blocks fused into standard convolutions at inference.
		(b) The features $\boldsymbol{F}^{32}$ undergo iterative self- and cross-attention, and are propagated to $1/8$ through RepFuse.
		(c) The interacted $\widetilde{\boldsymbol{F}}^{32}$ route candidate block pairs, whose corresponding $1/8$ features are indexed for sparse token matching. The resulting scores are assembled into a sparse global matrix $S$ and normalized by Sparse Global Dual-Softmax to obtain coarse correspondences $\mathcal{M}_c$.
		(d) For each coarse correspondence, indexed $\boldsymbol{F}^{f}$ and $\widetilde{\boldsymbol{F}}^{8}$ are fused and processed by shared lightweight encoders, followed by axis-wise heads that predict offset distributions and uncertainties for subpixel refinement.
	}
	\label{overview}
	\vspace{-10pt}
\end{figure}

As illustrated in Fig.~\ref{overview}, UltraMatch consists of four components. 
We first adopt structural reparameterization in the feature extractor 
(Sec.~\ref{Extraction}), followed by cross-image feature interaction 
(Sec.~\ref{Interaction}). We then perform Sparse Transport Path Matching 
to restrict token-level matching to routed candidates and avoid constructing 
the full matching matrix (Sec.~\ref{Matching}). Finally, a shared-parameter 
tiny fine matching head refines the coarse matches to subpixel accuracy 
(Sec.~\ref{FineMatching}).

\subsection{Structural Reparameterization for Feature Extraction}
\label{Extraction}

We first construct a highly compact feature extractor. 
Rather than further reducing network capacity, we adopt structural reparameterization~\citep{ding2019acnet,ding2021repvgg} to enrich the training-time representation while retaining the same compact single-branch convolutional structure at inference, since the parameter space does not necessarily coincide with the optimization space.
Specifically, reparameterizable convolutional blocks are employed from the $1/8$ to $1/32$ feature hierarchy. 
During training, the outputs of parallel \(3\times3\), \(1\times1\), \(1\times3\), and \(3\times1\) convolutional branches are summed and passed through a shared Batch Normalization layer. At inference, the padded branch kernels are summed, after which the shared normalization is folded into the equivalent \(3\times3\) convolution, introducing no additional branches in the deployed network, as shown in Fig.~\ref{overview}. The resulting $1/8$, $1/16$, and $1/32$ features are denoted as $\boldsymbol{F}_i^{8}$, $\boldsymbol{F}_i^{16}$, and $\boldsymbol{F}_i^{32}$ for image $I_i$, while an additional $1/8$ feature $\boldsymbol{F}_i^{f}$ is extracted for subsequent fine matching.

\subsection{Feature Interaction}
\label{Interaction}

To keep feature interaction efficient, contextual exchange is confined to the $1/32$ feature level, where a stack of $L$ alternating self-attention and cross-attention~\citep{vaswani2017attention} layers captures both intra-image context and inter-image correspondence cues. The interacted features $\widetilde{\boldsymbol{F}}_i^{32}$ are then propagated to the $1/16$ and $1/8$ levels through the Correlation Injection Module adopted from EDM, with its fusion convolutions reparameterized as described in Sec.~\ref{Extraction}. The resulting $\widetilde{\boldsymbol{F}}_i^{8}$ features are used for coarse matching, while $\widetilde{\boldsymbol{F}}_i^{32}$ features are retained for Transport Path Routing.

\subsection{Sparse Transport Path Matching}
\label{Matching}
Given the interacted features, we next perform coarse matching without constructing a dense similarity matrix over all token pairs. We apply routing directly to coarse matching, selecting which token similarities are evaluated and preserving global matching competition through geometry supervision, halo expansion, and sparse global Dual-Softmax. 
The interacted $1/32$ features already encode strong cross-image correspondence cues, making them a compact yet informative basis for identifying promising matching paths.

\paragraph{Transport Path Routing.}
Given the features $\widetilde{\boldsymbol{F}}_0^{32}$ and $\widetilde{\boldsymbol{F}}_1^{32}$, the routing features are obtained by
\begin{equation}
	\boldsymbol{D}_i =
	\operatorname{Norm}_2
	\left[
	\operatorname{Flatten}
	\left(
	\phi\left(
	\mathcal{A}(\widetilde{\boldsymbol{F}}_i^{32})
	\right)
	\right)
	\right],
	\quad i\in\{0,1\},
\end{equation}
where $\mathcal{A}(\cdot)$ aligns the interacted features with the routing grid, where each unit corresponds to a $B\times B$ block on the $1/8$ feature grid, and $\phi(\cdot)$ denotes a shared $1\times1$ projection. Let $\boldsymbol{d}_u^0$ and $\boldsymbol{d}_v^1$ denote the routing features of source block $u$ and target block $v$. Their affinity and the retained target blocks are defined jointly as
\begin{equation}
	S_{uv}^{r}
	=
	\frac{
		(\boldsymbol{d}_u^0)^{\mathsf T}\boldsymbol{d}_v^1
	}{
		\tau_r
	},
	\qquad
	\mathcal{N}_r(u)
	=
	\operatorname{TopR}(S_{u,:}^{r}),
\end{equation}
where $\mathcal{N}_r(u)$ specifies the candidate Transport Paths associated with source block $u$. As shown in Sec.~\ref{UnderstandingUltraMatch} and Appendix~\ref{High-resolutionMatching}, our Transport Path Routing strategy preserves 98.8\% ground-truth match coverage while accounting for less than 1\% of the total inference time on MegaDepth.

\paragraph{Sparse Token Matching.}
Each routed target block is expanded onto the $1/8$ feature grid. Let $\mathcal{T}_B(u)$ denote the set of $B\times B$ tokens associated with block $u$. To avoid discarding valid matches close to block boundaries, each routed target block is enlarged by a halo of $h$ tokens. The target candidate set of source block $u$ is therefore
\begin{equation}
	\mathcal{C}_u
	=
	\bigcup_{v\in\mathcal{N}_r(u)}
	\mathcal{H}_h
	\left(
	\mathcal{T}_B(v)
	\right),
\end{equation}
where $\mathcal{H}_h(\cdot)$ denotes halo expansion and overlapping candidates are merged. 
Token similarities are evaluated only within the routed candidate set:
\begin{equation}
	S_{ij}^{(u)}
	=
	\frac{
		\left(\widetilde{\boldsymbol{F}}_{0,i}^{8}\right)^{\mathsf T}
		\widetilde{\boldsymbol{F}}_{1,j}^{8}
	}{
		C\tau_c
	}
	+
	\omega S_{u,b(j)}^{r},
	\quad
	i\in\mathcal{T}_B(u),\;
	j\in\mathcal{C}_u,
\end{equation}
where $b(j)$ denotes the target block containing token $j$.
The coefficient is parameterized as
$\omega=\tanh(\widehat{\omega})$, where
$\widehat{\omega}$ is a learnable scalar initialized to zero.
Thus, the routing score not only determines the sparse candidate space
but also provides a block-level prior for token matching.

\paragraph{Sparse Global Dual-Softmax.}
Although the similarities above are computed locally for each source block, normalizing them independently would break the global competition critical to reliable matching. Instead, all routed token pairs are assembled into a sparse matching space
$\mathcal{G}=\bigcup_u \mathcal{T}_B(u)\times\mathcal{C}_u$,
with their similarities packed without constructing the complete matching matrix. Dual-Softmax is then applied directly over $\mathcal{G}$:
\begin{equation}
	P_{ij}
	=
	\frac{\exp(S_{ij})}
	{\sum_{j':(i,j')\in\mathcal{G}}\exp(S_{ij'})}
	\cdot
	\frac{\exp(S_{ij})}
	{\sum_{i':(i',j)\in\mathcal{G}}\exp(S_{i'j})},
	\qquad
	(i,j)\in\mathcal{G}.
\end{equation}
The row normalization competes among the routed candidates of each source token, while the column normalization gathers all routed connections arriving at the same target token across source blocks. The normalized scores are used to extract the coarse matches $\mathcal{M}_c$, preserving global competition without constructing the dense matching matrix. As shown in Sec.~\ref{UnderstandingUltraMatch}, this global normalization improves matching accuracy with negligible latency.

Since the routing overhead is negligible, as shown in Table~\ref{tab:runtime_breakdown}, we focus the complexity analysis on the token matching and normalization.
Let $N=H_cW_c$ be the number of tokens on the $1/8$ feature grid. Dense coarse matching requires $\mathcal{O}(N^2C)$ similarity computation, whereas UltraMatch restricts matching to $R$ routed $B\times B$ blocks, reducing it to
$
\mathcal{O}\left(NR(B+2h)^2C\right).
$
For fixed $h$, $R$ and $B$, the post-routing token matching complexity is therefore reduced from quadratic to approximately linear in $N$, since $N \gg R(B+2h)^2$ in practice. The Sparse Global Dual-Softmax is performed only over the routed edges, reducing its complexity from $\mathcal{O}(N^2)$ to $\mathcal{O}(|\mathcal{G}|)$, where $|\mathcal{G}| \ll N^2$.

\subsection{Shared Parameter Tiny Fine Matching}
\label{FineMatching}

To keep subpixel refinement lightweight, we design a compact fine matching head with extensive parameter sharing. The query and reference features are processed by the same residual encoder, avoiding separate branch-specific parameters while preserving symmetric feature processing. For each coarse match $(i,j)\in\mathcal{M}_c$, we index the corresponding features from $\boldsymbol{F}^{f}$ and the interacted $1/8$ feature $\widetilde{\boldsymbol{F}}^{8}$, and combine them by element-wise summation to form the query feature $q$ and reference feature $r$.
Instead of assigning separate encoders to the two roles, both features pass through the same residual encoder,
\begin{equation}
	q' = q + \mathcal{E}_d(q),
	\qquad
	r' = r + \mathcal{E}_d(r).
\end{equation}
The features are then concatenated in matching order and fused by a compact residual pair encoder,
\begin{equation}
	z=
	\mathrm{LN}\!\left(
	[q';r']
	+
	\mathcal{E}_p([q';r'])
	\right).
\end{equation}
For simplicity, normalization layers are absorbed into the corresponding encoder notation. Both residual encoders and the subsequent prediction heads are all implemented with simple two linear layers with GELU activation. Directional information is retained through the ordered concatenation, avoiding separate parameters for query and reference roles.
Two tiny axis heads then predict the horizontal and vertical discrete offset distributions together with their uncertainties. As in EDM~\citep{li2025edm}, the continuous displacement $\boldsymbol{\Delta}=(\Delta_x,\Delta_y)$ is recovered from the discrete bins by soft argmax, while $\boldsymbol{\sigma}=(\sigma_x,\sigma_y)$ denotes the corresponding uncertainties. The same fine matching head is applied to both $(q,r)$ and $(r,q)$ for bidirectional refinement.

\subsection{Loss Function}
\label{Loss}

\textbf{Transport Path Routing loss.}
The router is supervised by mapping ground-truth token correspondences to block pairs.
Let $\mathcal{U}$ denote source blocks with valid matches and $\mathcal{P}_u$ the corresponding positive target blocks. 
Since one source block may correspond to multiple targets, we adopt a multi-positive softmax loss:
\begin{equation}
	\mathcal{L}_{r}
	=
	-\frac{1}{|\mathcal{U}|}
	\sum_{u\in\mathcal{U}}
	\log
	\frac{
		\sum_{v\in\mathcal{P}_u}\exp(S^{r}_{uv})
	}{
		\sum_{v}\exp(S^{r}_{uv})
	}.
\end{equation}
To account for the halo used in sparse matching, $\mathcal{P}_u$ is further enlarged to a halo-aware set $\widehat{\mathcal{P}}_u$, yielding the coverage loss $\mathcal{L}_{\mathrm{cov}}$ with the same formulation.
Since these probability-based objectives do not explicitly guarantee that a valid path survives the routing cutoff, we further introduce a ranking loss.
For $(i,j)\in\mathcal{M}_{c}^{gt}$, let $s_{ij}$ be the best routing score among blocks covering $j$, and $\beta_{ij}$ the $R$-th largest routing score of the corresponding source block:
\begin{equation}
	\mathcal{L}_{\mathrm{rank}}
	=
	\frac{1}{|\mathcal{M}_{c}^{gt}|}
	\sum_{(i,j)\in\mathcal{M}_{c}^{gt}}
	\operatorname{softplus}
	\left(
	\beta_{ij}-s_{ij}+\mu
	\right),
\end{equation}
where $\mu$ is the ranking margin.
The complete routing objective is
$
\mathcal{L}_{\mathrm{route}}
=
\mathcal{L}_{r}
+
\mathcal{L}_{\mathrm{cov}}
+
\lambda_{\mathrm{rank}}\mathcal{L}_{\mathrm{rank}}.
$

\textbf{Coarse and Fine Matching Losses.}
For coarse matching, the focal loss used in ELoFTR is applied only to ground-truth matches contained in the routed matching space $\mathcal{G}$, denoted as $\mathcal{L}_{c}$. Fine refinement follows the RLE supervision adopted from EDM, denoted as $\mathcal{L}_{f}$.

The overall training objective is
\begin{equation}
	\mathcal{L}
	=
	\mathcal{L}_{c}
	+
	\lambda_{r}\mathcal{L}_{\mathrm{route}}
	+
	\lambda_{f}\mathcal{L}_{f}.
\end{equation}
\begin{figure}[t]
	\centering
	\begin{minipage}{0.32\linewidth}
		\centering
		\includegraphics[width=\linewidth]{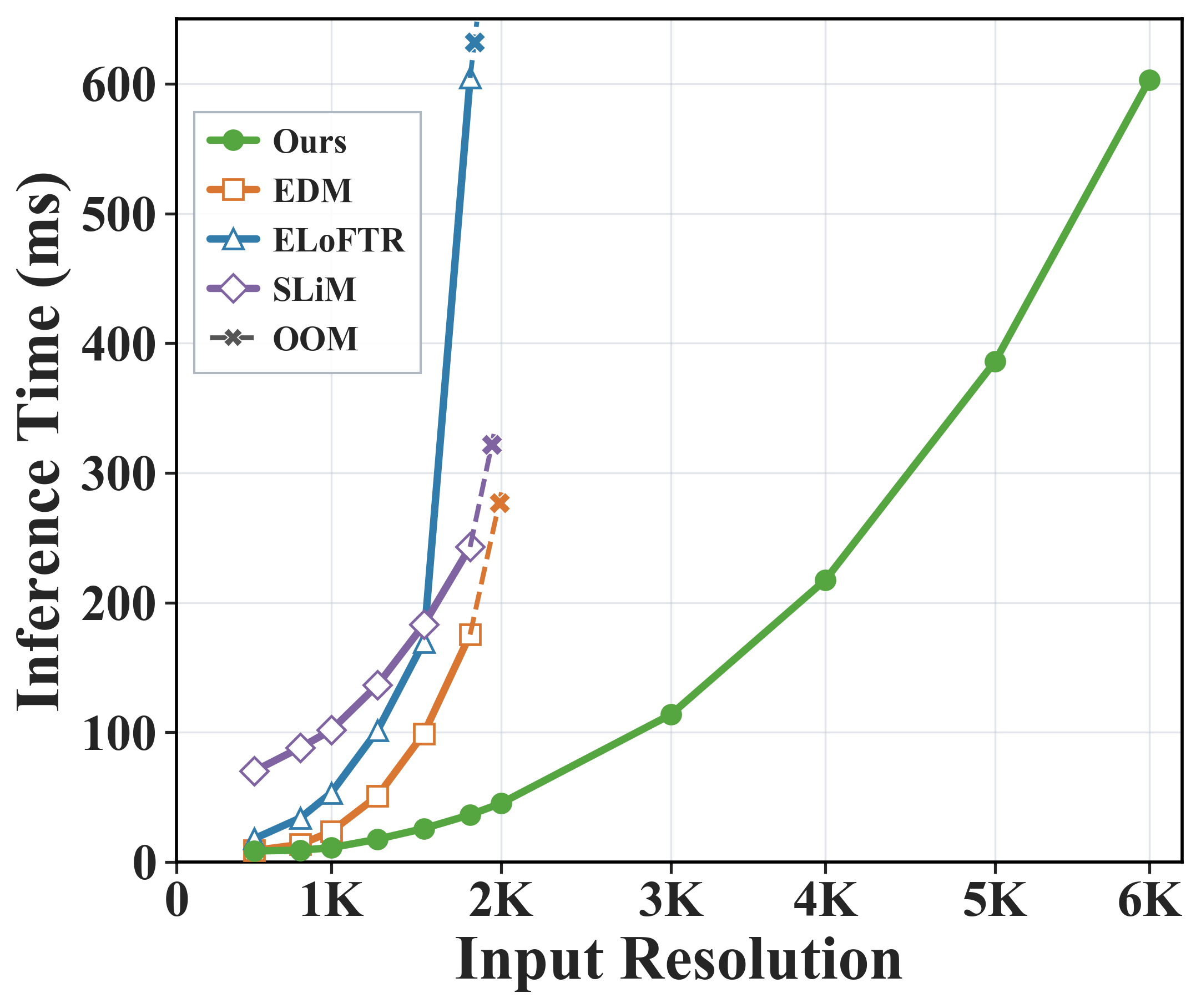}
		\small (a) Inference Runtime
	\end{minipage}
	\hfill
	\begin{minipage}{0.32\linewidth}
		\centering
		\includegraphics[width=\linewidth]{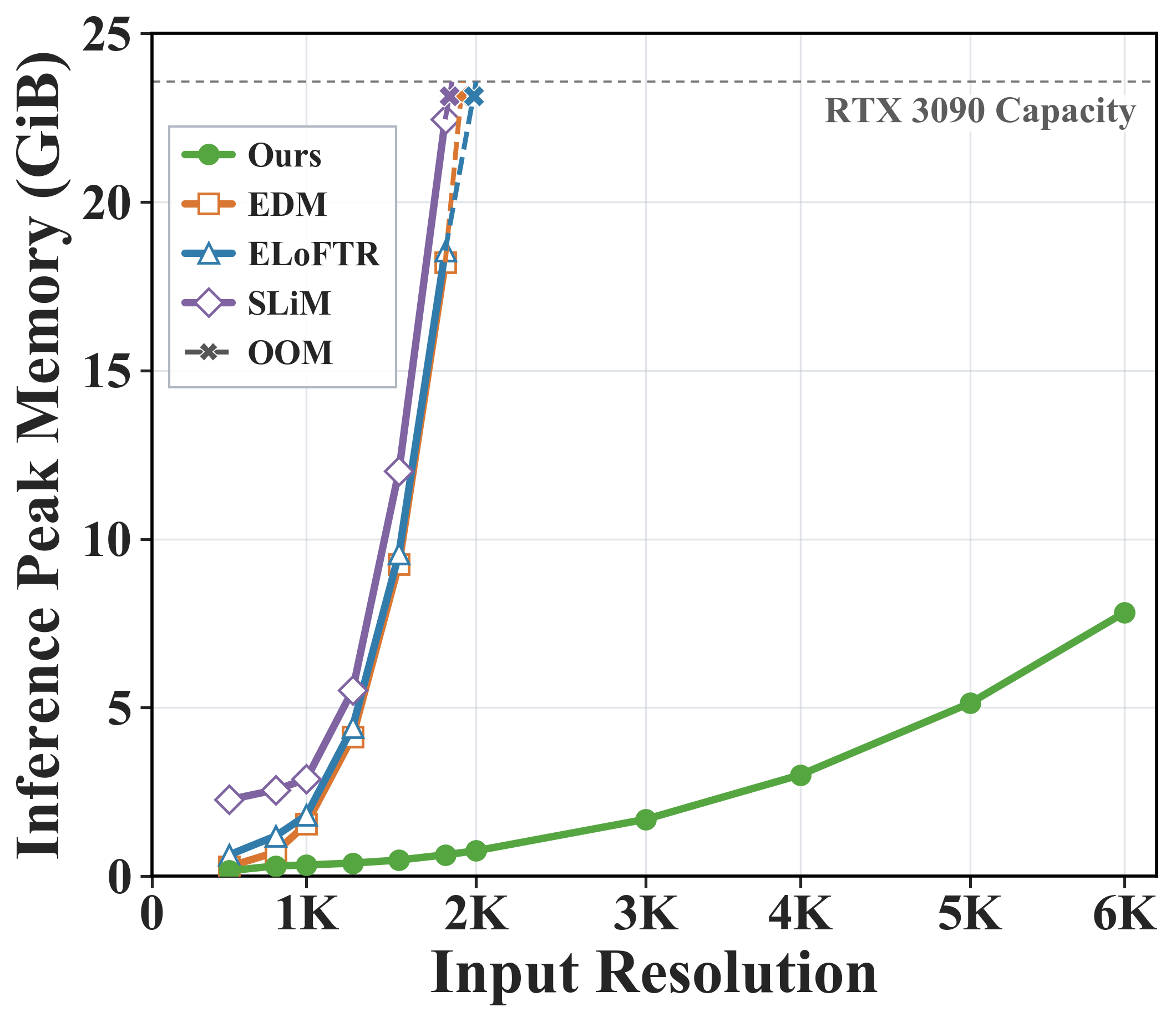}
		\small (b) Inference Peak Memory
	\end{minipage}
	\hfill
	\begin{minipage}{0.32\linewidth}
		\centering
		\includegraphics[width=\linewidth]{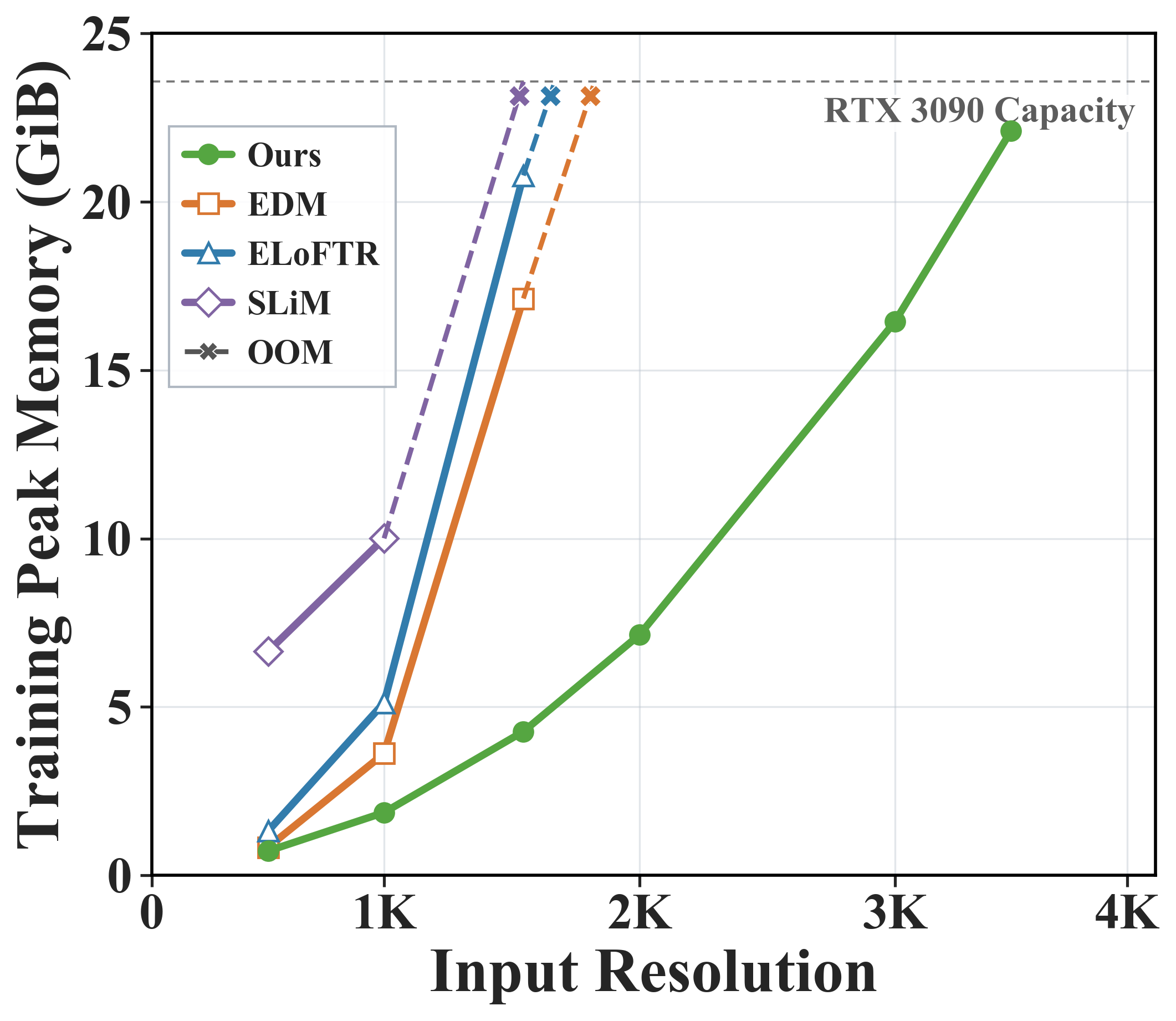}
		\small (c) Training Peak Memory
	\end{minipage}
	
	\caption{
		\textbf{Efficiency and scalability comparison under increasing input resolutions.}
		Inference runtime, peak inference memory, and peak training memory are reported.
		The horizontal dashed line indicates the effective memory capacity of an NVIDIA RTX 3090, while dashed curve extensions terminated by crosses denote out-of-memory (OOM) points.
	}
	\label{efficiency_scalability}
\end{figure}

\vspace{-10pt}
\subsection{Implementation Details}
\label{Implementation}

Feature interaction uses $L=2$ alternating self- and cross-attention blocks.
For Transport Path Routing, each routing unit corresponds to a $B=4$ block on the $1/8$ feature grid. The routing size cycles through $R\in\{4,6,8\}$ during training and is fixed to $R=6$ at inference, with a halo size of $h=1$.
The model is trained on MegaDepth~\citep{li2018megadepth} at $832\times832$ resolution for 30 epochs using AdamW with a total batch size of 32 across four RTX 3090 GPUs. Training completes in less than 7 hours. 
Further architectural and optimization details are provided in Appendix~\ref{MoreImplementationDetails}.

\begin{table}[t]
	\centering
	\caption{
		\textbf{Relative pose estimation on ScanNet and MegaDepth.}
		Pose AUCs at different thresholds are reported alongside runtime and peak GPU memory per image pair on MegaDepth. Bold and underlined values indicate the best and second-best results among semi-dense methods, respectively.
	}
	\vspace{-6pt}
	\label{main_results}
	
	\renewcommand{\arraystretch}{1.02}
	\setlength{\tabcolsep}{3.5pt}
	
	\resizebox{\linewidth}{!}{%
		\begin{tabular}{@{}llcccccccc@{}}
			\toprule
			
			\multirow{2}{*}{\textbf{Category}}
			& \multirow{2}{*}{\textbf{Method}}
			& \multicolumn{3}{c}{\textbf{ScanNet-1500} ($\uparrow$)}
			& \multicolumn{3}{c}{\textbf{MegaDepth-1500} ($\uparrow$)}
			& \multirow{2}{*}{\makecell{\textbf{Time}\\[-1pt]\textbf{(ms)($\downarrow$)}}}
			& \multirow{2}{*}{\makecell{\textbf{Memory}\\[-1pt]\textbf{(GiB)($\downarrow$)}}} \\
			
			\cmidrule(lr){3-5}
			\cmidrule(lr){6-8}
			
			&
			& AUC@$5^\circ$
			& AUC@$10^\circ$
			& AUC@$20^\circ$
			& AUC@$5^\circ$
			& AUC@$10^\circ$
			& AUC@$20^\circ$
			& & \\
			
			\midrule
			
			\multirow{2}{*}{Sparse}
			& SP + SG
			& 16.2
			& 32.8
			& 49.7
			& 49.7
			& 67.1
			& 80.6
			& 104.83
			& 1.02\\
			
			& SP + LG
			& 14.8
			& 30.8
			& 47.5
			& 49.9
			& 67.0
			& 80.1
			& 53.96
			& 1.02 \\
			
			\midrule
			
			\multirow{2}{*}{Dense}
			& DKM
			& 26.6
			& 47.1
			& 64.2
			& 60.4
			& 74.9
			& 85.1
			& 572.98
			& 9.66\\
			
			& RoMa
			& 28.9
			& 50.4
			& 68.3
			& 62.6
			& 76.7
			& 86.3
			& 755.44
			& 6.79 \\
			\midrule
			
			\multirow{9}{*}{Semi-Dense}
			& LoFTR
			& 16.9
			& 33.6
			& 50.6
			& 52.8
			& 69.2
			& 81.2
			& 376.15
			& 12.19 \\
			
			& QuadTree
			& 19.0
			& 37.3
			& 53.5
			& 54.6
			& 70.5
			& 82.2
			& 411.74
			& 12.20 \\
			
			& MatchFormer
			& 15.8
			& 32.0
			& 48.0
			& 53.3
			& 69.7
			& 81.8
			& 658.22
			& 8.58\\
			
			& ELoFTR
			& 19.2
			& 37.0
			& 53.6
			& 56.4
			& 72.2
			& 83.5
			& 140.49
			& 8.64\\
			
			& JamMa
			& 14.5
			& 29.8
			& 46.2
			& 55.4
			& 70.8
			& 82.1
			& 361.42
			& 5.49 \\
			
			& EDM
			& \underline{19.8}
			& \underline{37.5}
			& \underline{54.4}
			& \underline{57.5}
			& \textbf{73.2}
			& \textbf{84.2}
			& 83.45
			& 8.24\\

			& SLiM
			& 18.0
			& 34.7
			& 50.4
			& \textbf{57.9}
			& \underline{72.8}
			& 83.5
			& 157.63
			& 5.14 \\

			& UltraMatch (Ours)
			& \textbf{21.1}
			& \textbf{39.8}
			& \textbf{56.6}
			& 57.4
			& 72.5
			& \underline{83.6}
			& \textbf{32.26}
			& \textbf{0.44}\\

			\bottomrule
		\end{tabular}%
	}
	\vspace{-4.5mm}
\end{table}

\section{Experiments}

\subsection{Efficiency and Scalability Analysis}
\label{EfficiencyAnalysis}

\textbf{Datasets and Evaluation Protocol.}
We evaluate the inference and training scalability of UltraMatch over increasing input resolutions.
Inference experiments are conducted on ETH3D~\citep{schoeps2017multi}, whose native high-resolution images are directly resized to the target resolution without artificial padding, while training scalability is evaluated on MegaDepth.
We report inference latency, peak inference memory, and peak training memory, using a single NVIDIA RTX 3090 for inference and four RTX 3090 GPUs for training. A resolution is marked as OOM if any evaluation sample or the complete training step cannot be processed successfully.
The input sizes for each resolution level, full implementation and measurement details are provided in Appendix~\ref{ProtocolAppendix}.

\textbf{Results.}
As shown in Fig.~\ref{efficiency_scalability}, UltraMatch achieves substantial advantages in both efficiency and scalability over existing detector-free matchers. At $1824\times1216$, it requires only 36.43 ms and 0.63 GiB of inference memory, compared with 604.59 ms and 18.56 GiB for ELoFTR, yielding a $16.6\times$ speedup and a 96.6\% reduction in peak memory.
More importantly, this efficiency substantially expands the accessible resolution range of detector-free matching. Existing semi-dense methods under their official configurations exhaust GPU memory before or around the 2K regime during inference or training, whereas UltraMatch scales to $6048\times4032$ on a single RTX 3090 with only 7.82 GiB of inference memory. A similar advantage is observed during training, where UltraMatch supports substantially higher resolutions before reaching the GPU memory limit. These results demonstrate the practical scalability of UltraMatch for high-resolution and resource-constrained applications. Matching performance under increasing input resolutions is further evaluated in Appendix~\ref{High-resolutionMatching}.

\subsection{Relative Pose Estimation}
\label{sec:relative_pose}

\textbf{Datasets and Evaluation Protocol.}
We evaluate UltraMatch on MegaDepth-1500 and ScanNet-1500~\citep{dai2017scannet}, with UltraMatch trained exclusively on MegaDepth. For MegaDepth-1500, semi-dense methods are evaluated with images resized to \(1152\times1152\), while ScanNet-1500 images are resized to \(640\times480\).
Following the standard evaluation protocol~\citep{wang2024efficientloftr}, the pose error is defined as the maximum of the rotation and translation-direction errors, and we report AUC at $5^\circ$, $10^\circ$, and $20^\circ$. We additionally report inference latency and peak GPU memory measured on a single NVIDIA RTX 3090 GPU. Further protocol details are provided in Appendix~\ref{ProtocolAppendix}.

\textbf{Results.}
As shown in Tab.~\ref{main_results}, UltraMatch achieves strong relative pose estimation on both benchmarks. On ScanNet-1500, despite training only on MegaDepth, it achieves the best AUC among semi-dense methods at all three thresholds, showing strong cross-domain generalization. On MegaDepth-1500, it remains highly competitive with state-of-the-art matchers while offering substantial efficiency gains. UltraMatch requires only 32.26 ms and 0.44 GiB peak GPU memory, making it $4.35\times$ faster than ELoFTR and reducing peak memory by approximately 95\% under their official inference settings. Qualitative comparisons are provided in Appendix~\ref{QualitativeResults}.

\begin{table}[t]
	\centering
	\caption{
		\textbf{Evaluation of homography estimation on HPatches and visual localization on Aachen Day-Night v1.1 and InLoc.}
		Homography AUCs at 3, 5, and 10 pixels and percentages of correctly
		localized queries at three thresholds are reported.
	}
	\vspace{-6pt}
	\label{Homography}
	\small
	\renewcommand{\arraystretch}{1.1}
	\setlength{\tabcolsep}{3.5pt}
	
	\resizebox{\linewidth}{!}{%
		\begin{tabular}{@{}lccc@{\hspace{6pt}}cccc@{}}
			\toprule
			
			\multirow{3}{*}{\textbf{Method}}
			& \multicolumn{3}{c}{\textbf{HPatches} ($\uparrow$)}
			& \multicolumn{2}{c}{\textbf{Aachen Day-Night v1.1} ($\uparrow$)}
			& \multicolumn{2}{c}{\textbf{InLoc} ($\uparrow$)} \\
			
			\cmidrule(lr){2-4}
			\cmidrule(lr){5-6}
			\cmidrule(lr){7-8}
			
			& \multirow[c]{2}{*}{@3px}
			& \multirow[c]{2}{*}{@5px}
			& \multirow[c]{2}{*}{@10px}
			& \textbf{Day}
			& \textbf{Night}
			& \textbf{DUC1}
			& \textbf{DUC2} \\
			
			\cmidrule{5-8}
			
			& & & &
			\multicolumn{2}{c}{
				($0.25\,\mathrm{m},2^\circ$) /
				($0.5\,\mathrm{m},5^\circ$) /
				($5.0\,\mathrm{m},10^\circ$)
			} &
			\multicolumn{2}{c}{
				($0.25\,\mathrm{m},2^\circ$) /
				($0.5\,\mathrm{m},5^\circ$) /
				($1.0\,\mathrm{m},10^\circ$)
			} \\
			
			\midrule
			
			SP + SG
			& 37.0
			& 52.6
			& 70.1
			& 89.7 / 96.5 / 99.3
			& 73.8 / 91.1 / 99.5
			& 50.0 / 69.7 / 79.8
			& 47.3 / 77.9 / 80.2 \\
			
			SP + LG
			& 35.7
			& 51.6
			& 70.2
			& 89.2 / 96.5 / 99.3
			& 72.3 / 89.5 / 99.0
			& 48.0 / 68.7 / 79.8
			& 44.3 / 71.0 / 75.6 \\
			
			\midrule
			
			LoFTR
			& 51.3
			& 63.0
			& 75.5
			& 88.7 / \textbf{96.1} / 98.6
			& \textbf{77.0} / 90.6 / \textbf{99.5}
			& 49.0 / 71.7 / 84.3
			& 51.1 / 73.3 / 81.7 \\

			MatchFormer
			& 51.4
			& 63.1
			& 76.1
			& \textbf{89.4} / 96.0 / 98.8
			& 75.9 / 90.6 / \textbf{99.5}
			& 50.0 / 73.7 / 85.4
			& 58.0 / 80.9 / \textbf{87.0} \\
			
			ELoFTR
			& 53.5
			& 64.6
			& 75.9
			& 88.1 / 95.1 / 98.4
			& 73.8 / 90.6 / 98.4
			& \textbf{52.0} / 72.2 / 84.8
			& \textbf{59.5} / \textbf{82.4} / \textbf{87.0} \\
			
			JamMa
			& 49.9
			& 61.2
			& 74.0
			& 85.9 / 94.7 / 98.1
			& 72.8 / 90.1 / 97.9
			& 47.5 / 67.2 / 78.3
			& 35.9 / 53.4 / 69.5 \\
			
			EDM
			& 52.7
			& 64.7
			& 77.0
			& 87.4 / 95.6 / 98.2
			& 73.3 / 91.1 / 99.0
			& 47.5 / 70.2 / 81.3
			& 50.4 / 74.8 / 82.4 \\
			
			SLiM
			& 51.2
			& 62.4
			& 75.6
			& 71.8 / 79.4 / 85.7
			& 57.1 / 70.7 / 78.5
			& 40.4 / 58.1 / 68.2
			& 43.5 / 58.0 / 66.4 \\
			
			Ours
			& \textbf{54.0}
			& \textbf{65.4}
			& \textbf{77.5}
			& 87.9 / 95.4 / \textbf{99.0}
			& 75.4 / \textbf{91.6} / \textbf{99.5}
			& 50.0 / \textbf{74.2} / \textbf{85.9}
			& 55.0 / 76.3 / 81.7 \\
			
			\bottomrule
		\end{tabular}%
	}
	\vspace{-3mm}
\end{table}
\subsection{Homography Estimation}
\label{HomographyEstimation}
\textbf{Dataset and Evaluation Protocol.}
Homography estimation is evaluated on the HPatches~\citep{balntas2017hpatches} benchmark. All input images are resized such that the shorter side is 480 pixels, and the top 1,000 predicted matches are retained for evaluation. The homography is estimated with RANSAC, and its accuracy is measured by the mean reprojection error of the four image corners. Following ELoFTR~\citep{wang2024efficientloftr}, AUC is reported at thresholds of 3, 5, and 10 pixels.

\textbf{Results.}
As shown in Tab.~\ref{Homography},
UltraMatch consistently delivers the strongest homography estimation performance across all evaluated thresholds, demonstrating robust geometric accuracy on HPatches.

\subsection{Visual Localization}
\label{VisualLocalization}
\textbf{Datasets and Evaluation Protocol.}
Visual localization is evaluated on Aachen Day-Night v1.1~\citep{sattler2018benchmarking} and InLoc~\citep{taira2018inloc}, covering challenging outdoor and indoor scenarios, respectively. All methods are integrated into the HLoc~\citep{sarlin2019hloc} localization pipeline for a consistent evaluation. Performance is measured by the percentage of successfully localized queries under three pose-error thresholds. 

\textbf{Results.}
As reported in Tab.~\ref{Homography}, UltraMatch achieves competitive visual localization performance on both Aachen Day-Night v1.1 and InLoc, demonstrating robust generalization across challenging outdoor and indoor scenes.

\subsection{Understanding UltraMatch}
\label{UnderstandingUltraMatch}

\paragraph{Transferability of Transport Path Routing.}
Transport Path Routing is not specific to UltraMatch and can be readily transferred to existing semi-dense matchers to reduce computation and memory consumption. To examine this transferability, we integrate the complete routing strategy into ELoFTR, EDM, JamMa, and SLiM. Both the original baselines and their routed variants are trained from scratch following the official training protocols, with further details provided in Appendix~\ref{ProtocolAppendix}. As shown in Tab.~\ref{tab:router_transfer}, routing largely preserves matching accuracy while bringing substantial efficiency gains. ELoFTR, EDM, and JamMa achieve approximately $2\times$ end-to-end speedups, with substantially larger gains at the coarse matching stage. ELoFTR and EDM achieve around $10\times$ coarse-level acceleration, while JamMa reaches $29.02\times$. Peak GPU memory is also reduced by over 70\% in most cases and by 92.7\% for EDM. For JamMa, the end-to-end gain is partly offset by the increased fine-matching cost caused by its larger number of output matches.

\paragraph{Ablation Studies.}
All ablation variants are retrained under the same training protocol.
Tab.~\ref{tab:ablation} evaluates the main design choices of UltraMatch. 
Rows (1)–(2) show that structural reparameterization improves matching accuracy while preserving the same compact single-branch deployment form.
Replacing Transport Path Routing with dense 1/8 matching increases the runtime from 32.26 ms to 73.78 ms in row (3), demonstrating a 2.29$\times$ end-to-end acceleration attributable to routing within the same UltraMatch architecture.
Among the routing designs, row (5) replaces Sparse Global Dual-Softmax with independent Dual-Softmax normalization within each routed block, showing that cross-block competition is important for preserving matching accuracy. Halo expansion is likewise important, while multi-size training and the routing prior further improve robustness.
Finally, row (9) shows that sharing the query and reference encoders improves matching accuracy while avoiding duplicated encoder parameters. More experimental results are provided in the Appendix~\ref{MoreAblation}.
\begin{table*}[t]
	\centering
	
	\begin{minipage}[t]{0.595\textwidth}
		\centering
		\caption{
			\textbf{Transferability of Transport Path Routing.}
			Speedup and memory reduction are shown in parentheses.
		}
		\vspace{-6pt}
		\label{tab:router_transfer}
		\scriptsize
		\renewcommand{\arraystretch}{1.0}
		\setlength{\tabcolsep}{2.5pt}
		
		\resizebox{\linewidth}{!}{%
			\begin{tabular}{lccccc}
				\toprule
				\multirow{2}{*}{Method}
				& \multicolumn{2}{c}{Pose AUC}
				& \multirow{2}{*}{Time (ms)}
				& \multirow{2}{*}{Coarse (ms)}
				& \multirow{2}{*}{Mem. (GiB)} \\
				\cmidrule(lr){2-3}
				& @5$^\circ$ & @10$^\circ$ & & & \\
				\midrule
				
				ELoFTR
				& 54.9 & 71.2 & 141.61 & 77.88 & 8.64 \\
				+ Routing
				& \textbf{55.1} & \textbf{71.9}
				& 76.41 (\textbf{1.85$\times$})
				& 7.46 (\textbf{10.44$\times$})
				& 2.02 (\textbf{$-$76.6\%}) \\
				\midrule
				
				JamMa
				& 55.7 & 70.6 & 362.64 & 204.54 & 5.49 \\
				+ Routing
				& \textbf{56.0} & \textbf{71.3}
				& 187.05 (\textbf{1.94$\times$})
				& 7.05 (\textbf{29.02$\times$})
				& 1.43 (\textbf{$-$74.0\%}) \\
				\midrule
				
				SLiM
				& \textbf{55.4} & \textbf{69.2} & 193.30 & 59.72 & 5.14 \\
				+ Routing
				& 54.6 & 68.5
				& 147.30 (\textbf{1.31$\times$})
				& 14.22 (\textbf{4.20$\times$})
				& 4.06 (\textbf{$-$21.0\%}) \\
				\midrule
				EDM
				& 55.4 & 71.2 & 83.58 & 47.74 & 8.24 \\
				+ Routing
				& \textbf{56.1} & \textbf{72.1}
				& 41.56 (\textbf{2.01$\times$})
				& 4.73 (\textbf{10.09$\times$})
				& 0.60 (\textbf{$-$92.7\%}) \\
				\bottomrule
			\end{tabular}%
		}
	\end{minipage}
	\hfill
	\begin{minipage}[t]{0.39\textwidth}
		\centering
		\captionof{table}{
			\textbf{Ablation Studies on MegaDepth.}
		}
		\vspace{-6pt}
		\label{tab:ablation}
		\scriptsize
		\renewcommand{\arraystretch}{1.0}
		\setlength{\tabcolsep}{3.0pt}
		
		\resizebox{\linewidth}{!}{%
			\begin{tabular}{lccc}
				\toprule
				Variant
				& @5$^\circ$
				& @10$^\circ$
				& T. (ms) \\
				\midrule
				
				(1) w/o All Rep.
				& 55.6 & 71.7 & 35.12 \\
				
				(2) w/o Feature Rep.
				& 56.1 & 72.4 & 34.68 \\
				
				\midrule
				
				(3) Dense $1/8$ Matching
				& 55.8 & 71.9 & 73.78 \\
				
				(4) w/o Multi-Size
				& 56.0 & 71.7 & 32.35 \\
				
				(5) w/o Sparse Global DS
				& 54.1 & 69.7 & 32.68 \\
				
				(6) w/o Halo
				& 54.6 & 71.0 & \textbf{31.77} \\
				
				(7) Routing on $1/8$ Feat.
				& 56.4 & 72.1 & 39.56 \\
				
				(8) w/o Route Score Prior
				& 55.7 & 71.6 & 33.76 \\
				
				\midrule
				
				(9) w/o Shared Encoder
				& 55.0 & 71.5 & 35.68 \\
				
				\midrule
				
				\textbf{UltraMatch (Full)}
				& \textbf{57.4}
				& \textbf{72.5}
				& 32.26 \\
				
				\bottomrule
			\end{tabular}%
		}
	\end{minipage}
	\vspace{-12pt}
	
\end{table*}

\begin{wraptable}{r}{0.48\linewidth}
	\vspace{-6pt}
	\centering
	\caption{
		\textbf{Routing Size Analysis.}
	}
	\vspace{-8pt}
	\label{router_analysis}
	\small
	\renewcommand{\arraystretch}{1.0}
	\setlength{\tabcolsep}{2.5pt}
	
	\begin{tabular}{ccccc}
		\toprule
		$R$ & AUC@5$^\circ$ & T. (ms) & Mem. (GiB) & Rec. (\%) \\
		\midrule
		1     & 54.5 & \textbf{31.53} & \textbf{0.44} & 82.2 \\
		4     & 55.6 & 31.63 & \textbf{0.44} & 97.9 \\
		\textbf{6}
		& \textbf{57.4} & 32.26 & \textbf{0.44} & 98.8 \\
		8     & 56.6 & 33.87 & 0.45 & 99.2 \\
		Dense & 56.5 & 74.94 & 8.26 & \textbf{100.0} \\
		\bottomrule
	\end{tabular}
	
	\vspace{-6pt}
\end{wraptable}

\paragraph{Transport Path Routing Analysis.}
To evaluate whether the router preserves valid matching paths at inference, we vary the routing size $R$ using the same pretrained model. As shown in Tab.~\ref{router_analysis}, MegaDepth coverage increases from 82.2\% at $R=1$ to 98.8\% at the default $R=6$, showing that most candidate paths can be discarded while retaining nearly all ground-truth matches. On ScanNet, $R=6$ retains 88.7\%, further demonstrating the effectiveness of the routing strategy.

\section{Conclusions}

In this work, we presented UltraMatch, an ultra-efficient and scalable semi-dense matching framework that reduces dense matching cost through Transport Path Routing and sparse global Dual-Softmax. The routing strategy is readily transferable to existing semi-dense matchers, consistently reducing end-to-end latency with minimal accuracy change. Combined with structural reparameterization and a lightweight shared-parameter fine matching head, UltraMatch achieves competitive geometric accuracy with substantially lower latency and memory consumption, while scaling effectively to high-resolution inputs.

\subsection*{AI use statement}
In this work, generative AI tools were used for language polishing and grammar checking, as well as for assisting in diagnosing software issues during development. All AI-assisted revisions and diagnostic suggestions were carefully reviewed by the authors. Generative AI tools were not used to generate research ideas, design the proposed method or experiments, write or implement code, process or generate data, or interpret the experimental results. We take full responsibility for the final content of this work, including its text, code, claims, and experimental results.


\bibliography{iclr2027_conference}
\bibliographystyle{iclr2027_conference}

\clearpage
\appendix
\section{More Implementation Details}
\label{MoreImplementationDetails}

The feature dimensions at $1/8$, $1/16$, and $1/32$ resolutions are 128, 256, and 256, respectively, while both the fine feature $\boldsymbol{F}^{f}$ and the interacted $1/8$ feature $\widetilde{\boldsymbol{F}}^{8}$ are 256-dimensional.
The interacted $1/32$ features are projected from 256 to 64 dimensions using a shared $1\times1$ Conv--BN layer, followed by $\ell_2$ normalization. The routing descriptor dimension is $d_r=64$, with routing temperature $\tau_r=0.1$. We use a ranking margin of $\mu=0.5$ and a coarse-matching temperature of $\tau_c=0.1$. To obtain $\mathcal{M}_c$, each source token retains the highest-confidence target among its routed candidates, followed by global Top-$K$ selection. We set $(K,\theta_c)=(7257,0.10)$ for MegaDepth-1500 and $(1680,0.20)$ for ScanNet-1500.

The initial learning rate is $2\times10^{-3}$ with a weight decay of 0.1 and is halved at epochs $\{8,12,16,20,24\}$. We set $\lambda_{\mathrm{rank}}=0.5$, $\lambda_r=0.1$, and $\lambda_f=0.2$. In implementation, $\mathcal{L}_{r}$, $\mathcal{L}_{\mathrm{cov}}$,
and $\mathcal{L}_{\mathrm{rank}}$ are also computed in the reverse direction
by transposing $\boldsymbol{S}^{r}$ and exchanging the source and target
supervision, and the two directional losses are averaged.

The fine matching head operates on 256-dimensional features at $1/8$ resolution. For each spatial axis, it predicts 17 offset logits uniformly distributed over $[-0.5,0.5]$ together with one uncertainty logit. The continuous offset is recovered by soft argmax and scaled to the local resolution, corresponding to a displacement range of $[-4,4]$ pixels in the resized input image. The uncertainty logit is mapped by a sigmoid function to $\sigma_a\in(0,1)$, $a\in\{x,y\}$, and clamped to $[10^{-6},1-10^{-6}]$ during RLE training for numerical stability. The shared descriptor encoder, residual pair encoder, and axis-specific prediction heads use hidden dimensions of 256, 512, and 256, respectively. For each coarse correspondence, the same head predicts bidirectional refinements, with confidence defined as $1-(\sigma_x+\sigma_y)/2$. The higher-confidence direction is retained subject to the coarse confidence threshold and valid image boundaries.
\section{Evaluation Protocol Details}
\label{ProtocolAppendix}

\subsection{Efficiency Evaluation Protocol.}
Unless otherwise specified, all methods are evaluated using their official default inference configurations, including prescribed deployment optimizations and structural reparameterization. UltraMatch applies selective BF16 autocasting to feature extraction and interaction, while routing, coarse matching, and refinement remain in FP32. Runtime measures only the matcher forward pass, from preloaded GPU image tensors to the final refined correspondences. Image preprocessing, host-to-device transfer, and RANSAC are excluded. Latency is measured with CUDA events and explicit synchronization, while GPU memory is reported as the peak PyTorch allocated memory. Each method uses its official matching thresholds and native output settings without enforcing a common number of correspondences.

For Table~\ref{main_results}, we perform 50 warm-up forwards and report the mean latency over all 1,500 MegaDepth pairs. For the scalability experiments in Fig.~\ref{efficiency_scalability}, we perform 10 warm-up forwards at each resolution and report the mean latency on ETH3D. Training memory is evaluated with a local batch size of 1 and a global batch size of 4 across four RTX 3090 GPUs. 

The input sizes corresponding to each resolution level are summarized in Table~\ref{resolution_settings}, where ``K'' approximately denotes the number of pixels along the image's longer side.

\begin{table}[t]
	\centering
	\caption{\textbf{Input-resolution settings used in the high-resolution experiments.}
		Each resolution denotes the size of an individual input image.}
	\label{resolution_settings}
	\small
	\renewcommand{\arraystretch}{1.05}
	\setlength{\tabcolsep}{7pt}
	\begin{tabular}{lc}
		\toprule
		Resolution level & Input size ($W \times H$) \\
		\midrule
		0.5K  & $480 \times 320$   \\
		0.75K & $768 \times 512$   \\
		1K    & $960 \times 640$   \\
		1.25K & $1248 \times 832$  \\
		1.5K  & $1536 \times 1024$ \\
		1.8K  & $1824 \times 1216$ \\
		2K    & $2016 \times 1344$ \\
		3K    & $3072 \times 2048$ \\
		4K    & $4032 \times 2688$ \\
		5K    & $5088 \times 3392$ \\
		6K    & $6048 \times 4032$ \\
		\bottomrule
	\end{tabular}
\end{table}

\subsection{Relative Pose Estimation.}
For MegaDepth-1500, semi-dense methods are evaluated with images resized to $1152\times1152$, while ScanNet-1500 images are resized to $640\times480$. Relative pose is estimated from the refined correspondences using OpenCV RANSAC with an essential-matrix model. The inlier threshold and confidence are set to $0.5$ pixels and $0.99999$, respectively, with a minimum of five correspondences. If multiple essential matrices are returned, the solution producing the largest number of pose inliers after \texttt{recoverPose} is selected.  SuperPoint extracts up to 2,048 keypoints per image, using a detection threshold of 0.005 and an NMS radius of 3. LightGlue uses adaptive early stopping and point pruning with a match filtering threshold of 0.1. 

The pose error is defined as the maximum of the rotation error and translation-direction error, following the standard evaluation protocol~\citep{wang2024efficientloftr}. We report the area under the cumulative pose error curve (AUC) at $5^\circ$, $10^\circ$, and $20^\circ$.

\subsection{Transferability Experimental Details.}
Transport Path Routing is integrated into ELoFTR, JamMa, SLiM, and EDM only at the coarse matching stage, leaving their feature extraction, interaction, and fine refinement modules unchanged. Each model’s final $1/8$ coarse features are average-pooled over $4\times4$ blocks and projected to 64-dimensional routing descriptors. Matching and normalization are restricted to the routed candidates while retaining architecture-specific probability formulations and filtering rules. All variants use $\tau_r=0.1$, $R=6$, and a one-token halo, with training budgets $R\in{4,6,8}$ and a routing loss weight of $0.1$. Both baselines and routed variants are trained from scratch at $832\times832$ using each architecture’s optimizer and schedule. ELoFTR, JamMa, SLiM, and EDM use global batch sizes of 8, 16, 4, and 32, respectively, for 30 epochs.

\section{High-resolution matching analysis.}
\label{High-resolutionMatching}

To examine whether the scalability of UltraMatch translates into reliable high-resolution matching, we evaluate downstream relative pose estimation, routing coverage, and routing overhead on all 13 scenes of the ETH3D high-resolution training split, using six fixed image pairs per scene. Each full-frame image is directly resized, without padding, from $480\times320$ up to $6048\times4032$. Ground-truth correspondences are generated using the provided laser depth maps, calibrated cameras, and camera poses. We report scene-macro results to give each scene equal weight.

\begin{figure}[t]
	\centering
	\begin{minipage}{0.325\linewidth}
		\centering
		\includegraphics[width=\linewidth]{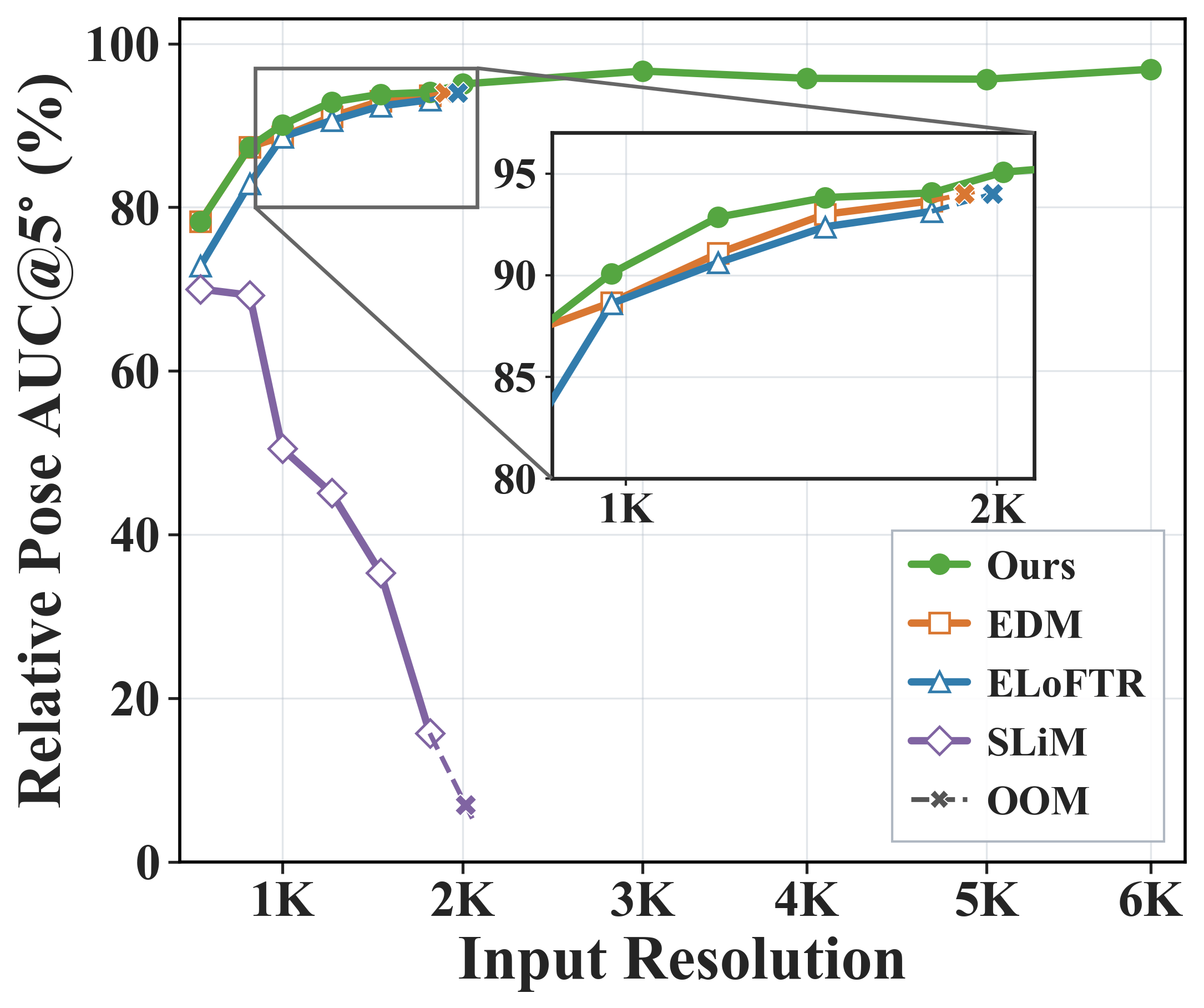}
		\small (a) Relative Pose AUC@5$^\circ$
	\end{minipage}
	\hfill
	\begin{minipage}{0.325\linewidth}
		\centering
		\includegraphics[width=\linewidth]{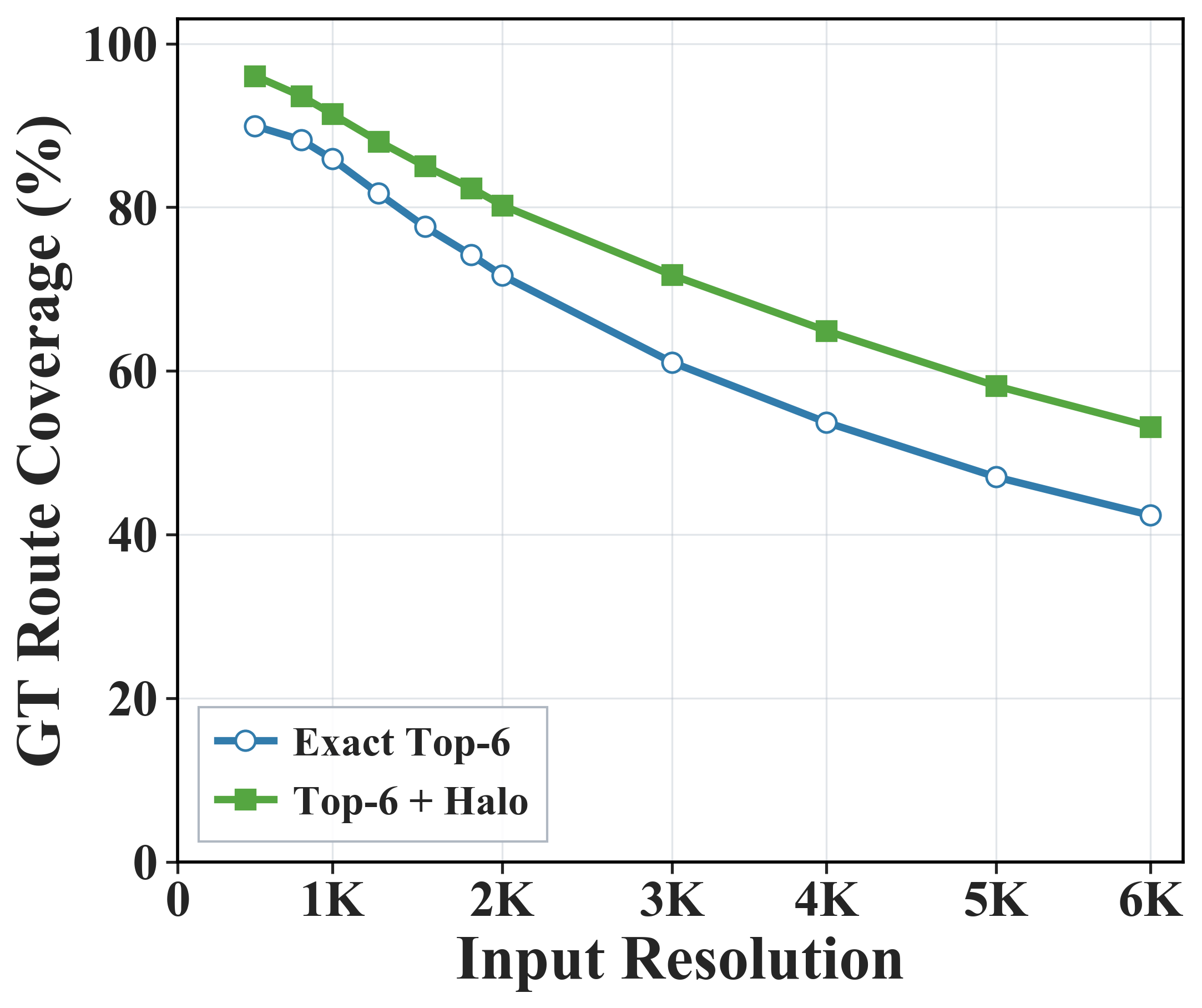}
		\small (b) GT Route Coverage
	\end{minipage}
	\hfill
	\begin{minipage}{0.325\linewidth}
		\centering
		\includegraphics[width=\linewidth]{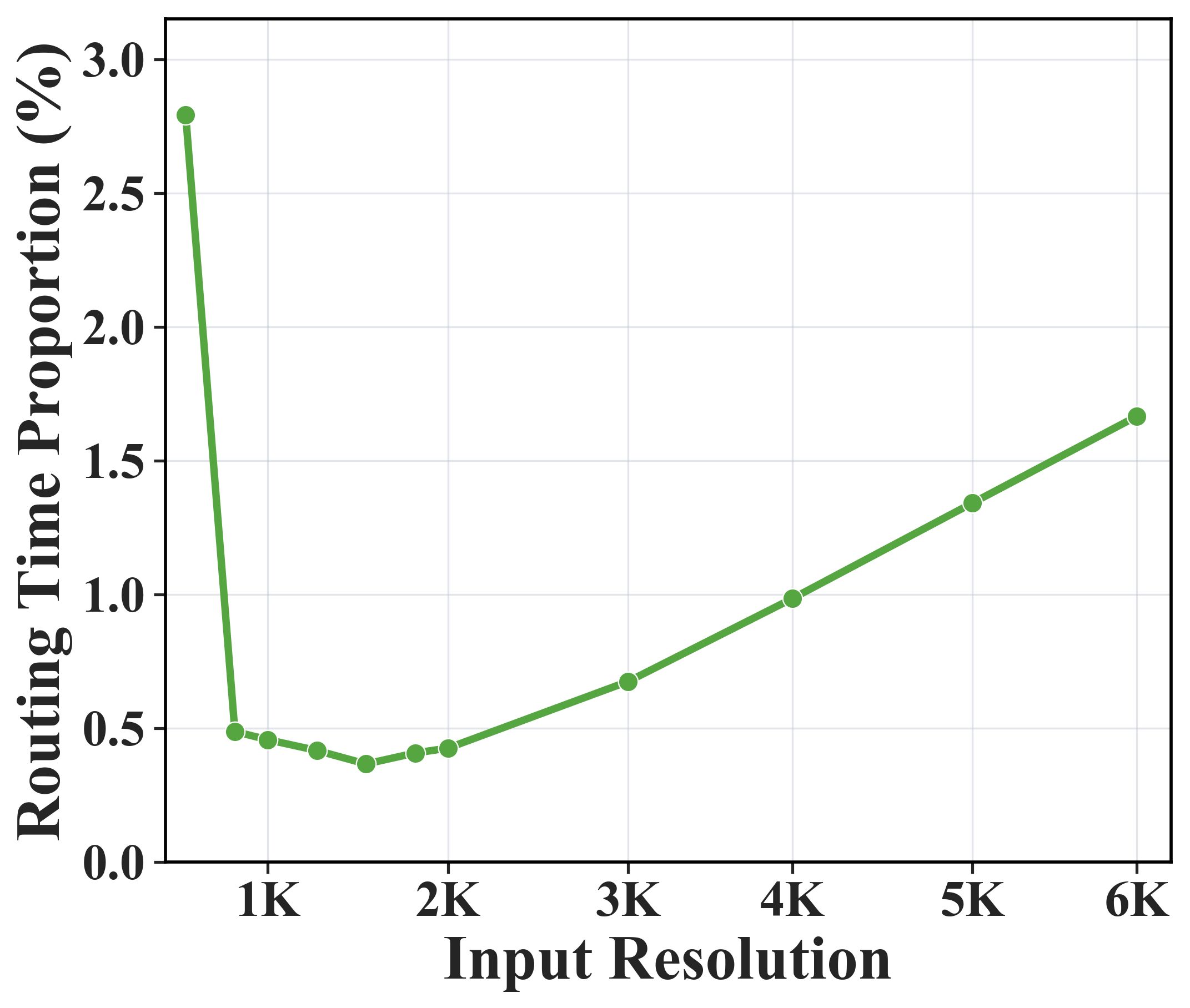}
		\small (c) Routing Time Proportion
	\end{minipage}
	
	\caption{
		\textbf{High-resolution matching analysis on ETH3D.}
		(a) Relative pose AUC@5$^\circ$ under increasing input resolutions.
		UltraMatch maintains strong geometric accuracy up to 6K resolution, whereas competing semi-dense matchers either reach their memory limits at substantially lower resolutions or suffer pronounced accuracy degradation.
		(b) Ground-truth route coverage, comparing the exact Top-6 routed blocks with the candidate space after halo expansion.
		Although GT coverage decreases as the token space grows, halo expansion consistently recovers additional valid correspondence paths.
		(c) Proportion of total inference time spent on routing score computation, which remains below 3\% across all evaluated resolutions.
	}
	\label{fig:high_resolution_analysis}
\end{figure}

As shown in Fig.~\ref{fig:high_resolution_analysis}(a), the relative pose AUC@5$^\circ$ of UltraMatch improves up to approximately 2K--3K and then remains stable through 6K. In contrast, EDM and ELoFTR encounter OOM at 2K on a 24\,GB RTX~3090, while SLiM exhibits substantial accuracy degradation before also encountering OOM during the 2K evaluation. This confirms that the scalability of UltraMatch translates into reliable geometric estimation rather than merely enabling higher-resolution inference.

Fig.~\ref{fig:high_resolution_analysis}(b) further shows that the GT coverage of the exact Top-6 routed blocks gradually decreases as resolution increases. Higher resolutions rapidly increase both token and block counts, so a fixed Top-$R$ size covers a smaller fraction of candidate blocks, making it more challenging to retain all valid correspondence paths. Nevertheless, halo expansion consistently recovers additional valid correspondence paths, and the retained candidates remain sufficiently informative for accurate pose estimation.

Finally, Fig.~\ref{fig:high_resolution_analysis}(c) reports the proportion of inference time spent on routing score computation. Although this proportion increases with resolution, it remains below 3\% throughout the evaluated range and is 1.66\% at 6K, confirming that routing itself introduces only a minor practical overhead.

\section{Qualitative Results}
\label{QualitativeResults}
\begin{figure}[t]
	\centering
	\includegraphics[width=\linewidth]{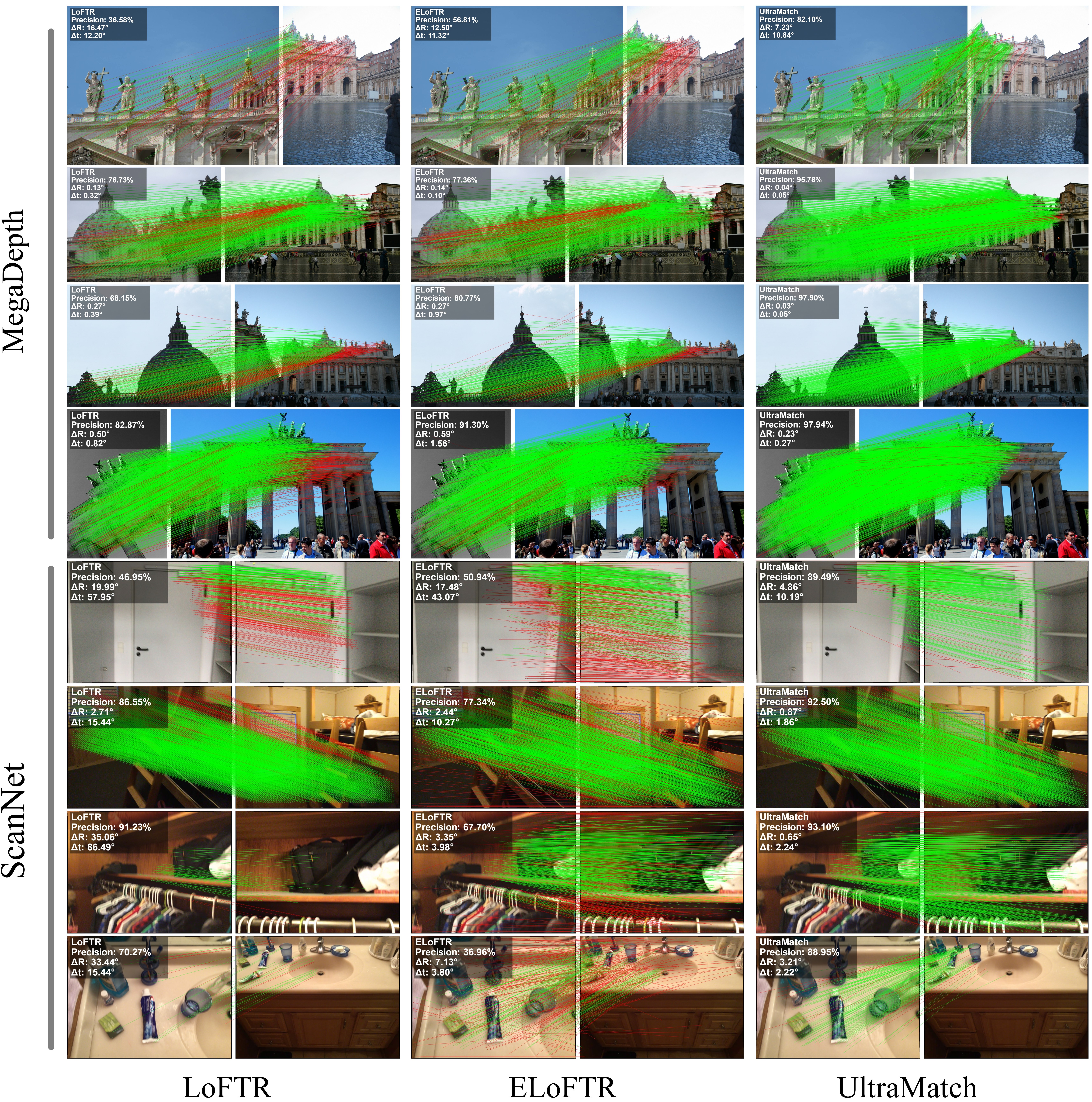}
	\caption{\textbf{Qualitative matching comparisons of UltraMatch, LoFTR, and ELoFTR on MegaDepth and ScanNet.} Green lines indicate geometrically consistent matches, while red lines denote matches whose epipolar error exceeds \(5\times10^{-4}\) in normalized image coordinates.
	}
	\label{qualitative_results}
\end{figure}

As shown in Fig.~\ref{qualitative_results}, UltraMatch produces dense and geometrically consistent matches across challenging cases involving large scale variations, weak textures, and illumination changes. Despite its substantial advantage in computational efficiency, the matching quality remains highly competitive with LoFTR and ELoFTR, further demonstrating the robustness of UltraMatch in challenging  scenes.

\section{Runtime Breakdown.}
\label{RuntimeBreakdown}
We report the stage-wise runtime of UltraMatch in Tab.~\ref{tab:runtime_breakdown}. Most computation is spent on feature extraction and feature interaction, while the router introduces only 0.32 ms of additional overhead. Matching and refinement take 4.27 ms and 4.17 ms, respectively, resulting in a total runtime of 32.26 ms. This breakdown further confirms that Transport Path Routing substantially reduces matching cost with negligible routing overhead.

\begin{table}[t]
	\centering
	\caption{
		\textbf{Runtime breakdown of UltraMatch.}
	}
	\label{tab:runtime_breakdown}
	\small
	\renewcommand{\arraystretch}{1.0}
	\setlength{\tabcolsep}{8.0pt}
	
	\begin{tabular}{lcc}
		\toprule
		\textbf{Stage} 
		& \textbf{Time (ms)}
		& \textbf{Ratio (\%)} \\
		\midrule
		Feature Extraction  & 10.80 & 33.48 \\
		Feature Interaction & 12.13 & 37.60 \\
		Router              & 0.32  & 0.99 \\
		Coarse Matching            & 4.27  & 13.24 \\
		Refinement          & 4.17  & 12.93 \\
		Other Overhead      & 0.57  & 1.77 \\
		\midrule
		\textbf{Total}      
		& \textbf{32.26}
		& \textbf{100.00} \\
		\bottomrule
	\end{tabular}
	
\end{table}

\section{More Ablation Studies.}
\label{MoreAblation}
\subsection{Ablation of Routing Losses.}
Tab.~\ref{tab:loss_ablation} evaluates the contribution of the three routing objectives. 
The basic routing loss $\mathcal{L}_{r}$ already improves matching performance, while the coverage-aware loss $\mathcal{L}_{\mathrm{cov}}$ brings a further clear gain by encouraging valid correspondences to remain within the routed candidate space. 
The ranking loss $\mathcal{L}_{\mathrm{rank}}$ provides an additional improvement by explicitly promoting valid paths above the Top-$R$ selection boundary. 

\begin{table}[t]
	\centering
	\caption{
		\textbf{Ablation study of routing losses on MegaDepth.}
	}
	\label{tab:loss_ablation}
	\scriptsize
	\renewcommand{\arraystretch}{1.05}
	\setlength{\tabcolsep}{5.0pt}
	
	\begin{tabular}{ccc|ccc}
		\toprule
		$\mathcal{L}_{r}$
		& $\mathcal{L}_{\mathrm{cov}}$
		& $\mathcal{L}_{\mathrm{rank}}$
		& @5$^\circ$
		& @10$^\circ$
		& T. (ms) \\
		\midrule
		
		& & 
		& 52.4 & 68.5 & 32.12 \\
		
		\checkmark & &
		& 55.7 & 72.0 & 32.88 \\
		
		\checkmark & \checkmark &
		& 56.6 & 72.5 & 32.48 \\
		
		\checkmark & \checkmark & \checkmark
		& \textbf{57.4}
		& \textbf{72.5}
		& 32.26 \\
		
		\bottomrule
	\end{tabular}
\end{table}

\subsection{Memory-Efficient Dense Matching.}
To disentangle the benefit of matching sparsification from that of avoiding full-matrix materialization, we compare native dense matching, a streaming dense implementation, and Transport Path Routing in Tab.~\ref{tab:memory_analysis}. Native dense matching incurs substantial memory consumption due to the full similarity matrix. Streaming Dual-Softmax avoids materializing this matrix and reduces peak memory from 8.26\,GiB to 0.43\,GiB, but still exhaustively traverses the complete similarity space and requires recomputation, increasing the matching time to 289.22\,ms. In contrast, Transport Path Routing achieves similarly low memory consumption while reducing the matching time to 32.26\,ms by directly sparsifying the matching search space. This shows that the main benefit of routing comes from reducing the underlying matching computation rather than merely avoiding dense matrix materialization. The streaming baseline uses a PyTorch chunked implementation with $1024\times1024$ tiles and traverses the complete similarity space twice to compute exact global Dual-Softmax without materializing the full confidence matrix.
\begin{table}[t]
	\centering
	\caption{
		\textbf{Analysis of matching sparsification and memory-efficient normalization on MegaDepth.}
		Relative pose AUC (\%), average pairwise matching time, and peak allocated GPU memory at $1152\times1152$ resolution are reported.
	}
	\label{tab:memory_analysis}
	\small
	\renewcommand{\arraystretch}{1.05}
	\setlength{\tabcolsep}{6pt}
	\begin{tabular}{lccccc}
		\toprule
		\textbf{Variant} &
		\textbf{AUC@5$^\circ$ $\uparrow$} &
		\textbf{AUC@10$^\circ$ $\uparrow$} &
		\textbf{Time (ms) $\downarrow$} &
		\textbf{Memory (GiB) $\downarrow$} \\
		\midrule
		Dense Dual-Softmax (Native)
		& 56.5 & 72.4 
		& 74.94 & 8.26 \\
		
		Dense Dual-Softmax (Streaming)
		& 56.5 & 72.4 
		& 289.22 & \textbf{0.43} \\
		
		\textbf{Transport Path Routing}
		& \textbf{57.4} & \textbf{72.5} 
		& \textbf{32.26} & 0.44 \\
		\bottomrule
	\end{tabular}
\end{table}

\subsection{Controlled Efficiency Comparison.}
Tab.~\ref{tab:precision_fairness} further compares efficiency under explicitly specified inference configurations. For SuperPoint+LightGlue, $K$ denotes the maximum number of
keypoints per image.
``Adaptive'' denotes enabled early stopping and point pruning, with a match filtering
threshold of 0.1.
FlashAttention is enabled in the configuration, without
\texttt{torch.compile}.
``FP32 + FP16 attn.'' denotes FP32 inference with global
mixed precision disabled, while Q/K/V
are internally cast to FP16 in the Flash attention path.
Runtime includes SuperPoint extraction for both images
and LightGlue matching.
UltraMatch outputs 3973 correspondences per pair on average, more than both ELoFTR-Full (3288) and ELoFTR-Opt (3531), and comparable to EDM (4326), indicating that its efficiency advantage is not obtained by aggressively reducing the correspondence output.
Even under full FP32 inference, UltraMatch requires only 37.85\,ms and 0.57\,GiB, substantially lower than ELoFTR-Full, its optimized configuration ELoFTR-Opt, and EDM.
Selective BF16 further reduces the latency to 32.26\,ms and memory to 0.44\,GiB, corresponding to only a $1.17\times$ additional speedup over FP32 while preserving the pose accuracy.
Together with the routing ablation in Tab.~\ref{tab:ablation}, these results show that numerical precision provides only a limited additional gain, while the primary efficiency improvement stems from Transport Path Routing. For ELoFTR-Opt, $\theta_c$ is applied to the raw scaled similarity logits
because Dual-Softmax is skipped. So it is not numerically comparable
to the confidence threshold used by ELoFTR-Full.

\begin{table}[t]
	\centering
	\caption{
		\textbf{Controlled efficiency comparison and inference configurations on MegaDepth-1500.}
		Accuracy, correspondence count, runtime, and memory are measured under the listed inference configuration. \textsuperscript{\dag}
		``Adaptive, $K\leq2048$'' denotes at most 2048 keypoints per
		image, with early stopping
		and point pruning enabled.
		``FP32 + FP16 attn.'' denotes FP32 inference with Q/K/V
		internally cast to FP16 in the Flash attention path.
	}
	\label{tab:precision_fairness}
	\small
	\renewcommand{\arraystretch}{1.05}
	\setlength{\tabcolsep}{4.0pt}
	\resizebox{\linewidth}{!}{
		\begin{tabular}{lcccccccc}
			\toprule
			\textbf{Method} &
			\textbf{Input} &
			\textbf{Configuration} &
			\textbf{Precision} &
			\textbf{AUC@5$^\circ$} &
			\textbf{\# Matches} &
			\textbf{Time (ms)} &
			\textbf{Memory (GiB)} \\
			\midrule
			
			SuperPoint+LightGlue & $1152^2$
			& Adaptive, $K\leq2048$\,\textsuperscript{\dag}
			& FP32 + FP16 attn.\,\textsuperscript{\dag}
			& 49.7 & 553 & 53.96 & 1.02 \\
			ELoFTR-Full
			& $1152^2$
			& Full, $\theta_c{=}0.1$
			& Mixed FP16
			& 56.4 & 3288 & 140.49 & 8.64 \\
			
			ELoFTR-Opt
			& $1152^2$
			& Opt, $\theta_c{=}20$
			& Mixed FP16
			& 55.4 & 3531 & 92.17 & 3.43 \\
			
			EDM
			& $1152^2$
			& $\theta_c{=}0.05$
			& FP32
			& \textbf{57.5} & 4326 & 83.45 & 8.24 \\
			
			\midrule
			
			UltraMatch
			& $1152^2$
			& $R{=}6$, halo${=}1$, $\theta_c{=}0.10$
			& FP32
			& 57.4 & 3973 & 37.85 & 0.57 \\
			
			UltraMatch
			& $1152^2$
			& $R{=}6$, halo${=}1$, $\theta_c{=}0.10$
			& Selective BF16
			& 57.4 & 3973 & \textbf{32.26} & \textbf{0.44} \\
			\bottomrule
		\end{tabular}
	}
\end{table}

\subsection{Future Directions}
\label{discussion}

UltraMatch uses a fixed routing size of $R=6$ across all input resolutions. As resolution increases, each source block still retains only six paths over an expanding candidate space, making the observed decrease in correspondence coverage expected. Importantly, our high-resolution experiments show that this does not translate into degraded pose estimation, indicating that the retained correspondences remain sufficiently informative for the evaluated task. Nevertheless, tasks requiring more extensive spatial coverage may benefit from dynamically adapting the routing size to resolution, routing uncertainty, or matching ambiguity.

Tab.~\ref{tab:runtime_breakdown} shows that feature extraction and interaction account for 71.1\% of the total runtime, while routing itself contributes less than 1\%. Their dominant share mainly reflects the extremely low cost of the subsequent matching stages. Our future research will build on this highly efficient architecture and further compress feature extraction and interaction without sacrificing geometric accuracy, aiming to push semi-dense matching toward the practical limits of latency, memory, and accuracy.

\end{document}